\documentclass[11pt]{article}
\usepackage{coling2018}
\usepackage{times}
\usepackage{url}
\usepackage{latexsym}
\usepackage{graphicx}
\usepackage{url}
\usepackage{multirow}
\usepackage{amsmath}
\usepackage{amsfonts,amssymb}
\usepackage{subfigure}
\usepackage{float}
\newfloat{figtab}{htb}{fgtb}
\makeatletter
\newcommand\figcaption{\def\@captype{figure}\caption}
\newcommand\tabcaption{\def\@captype{table}\caption}
\makeatother
\title{Lingke: A Fine-grained Multi-turn Chatbot for Customer Service}

\author{Pengfei Zhu$^{1,2,3}$, Zhuosheng Zhang$^{2,3}$, Jiangtong Li$^{2,3,4}$, Yafang Huang$^{2,3}$, Hai Zhao$^{2,3,\dagger}$\\
	$^{1}$School of Computer Science and Software Engineering, East China Normal University, China\\
	$^{2}$Department of Computer Science and Engineering, \\Shanghai Jiao Tong University, China \\
	$^{3}$Key Laboratory of Shanghai Education Commission for Intelligent Interaction \\ and Cognitive Engineering, Shanghai Jiao Tong University, Shanghai, 200240, China\\
	$^{4}$College of Zhiyuan, Shanghai Jiao Tong University, China\\
	{\tt 10152510190@stu.ecnu.edu.cn, \{zhangzs, keep\_moving-lee\}@sjtu.edu.cn,  } \\ {\tt huangyafang@sjtu.edu.cn, zhaohai@cs.sjtu.edu.cn}
}

\date{}

\begin{document}
\maketitle
\begin{abstract}
	Traditional chatbots usually need a mass of human dialogue data, especially when using supervised machine learning method. Though they can easily deal with single-turn question answering, for multi-turn the performance is usually unsatisfactory. In this paper, we present Lingke, an information retrieval augmented chatbot which is able to answer questions based on given product introduction document and deal with multi-turn conversations. We will introduce a fine-grained pipeline processing to distill responses based on unstructured documents, and attentive sequential context-response matching for multi-turn conversations.
\end{abstract}

\section{Introduction}

\blfootnote{
	$\dagger$ Corresponding author. This paper was partially supported by National Key Research and Development Program of China (No. 2017YFB0304100),
	National Natural Science Foundation of China (No. 61672343 and No. 61733011),
	Key Project of National Society Science Foundation of China (No. 15-ZDA041),
	The Art and Science Interdisciplinary Funds of Shanghai Jiao Tong University (No. 14JCRZ04).
}
\blfootnote{
	This work is licensed under a Creative Commons Attribution 4.0 International Licence. Licence details: \url{http://creativecommons.org/licenses/by/4.0/}.
}

Recently, dialogue and interactive systems have been emerging with huge commercial values \cite{Qiu2017AliMe,Yan2016DocChat,Zhang2017Benben,Huang2018Moon,zhang2018dua,zhang2018SubMRC}, especially in  the e-commerce field \cite{Cui2017SuperAgent,Yan2017Building}. Building a chatbot mainly faces two challenges, the lack of dialogue data and poor performance for multi-turn conversations. This paper describes a fine-grained information retrieval (IR) augmented multi-turn chatbot - Lingke. It can learn knowledge without human supervision from conversation records or given product introduction documents and generate proper response, which alleviates the problem of lacking dialogue corpus to train a chatbot. First, by using \emph{Apache Lucene}\footnote{\url{http://lucene.apache.org}} to select top 2 sentences most relevant to the question and extracting subject-verb-object (SVO) triples from them, a set of candidate responses is generated. With regard to multi-turn conversations, we adopt a dialogue manager, including self-attention strategy to distill significant signal of utterances, and sequential utterance-response matching to connect responses with conversation utterances, which outperforms all other models in multi-turn response selection. An online demo is available via accessing \url{http://47.96.2.5:8080/ServiceBot/demo/}.

\section{Architecture}
\begin{figure*}[htb]
	\centering
	\includegraphics[width=\textwidth]{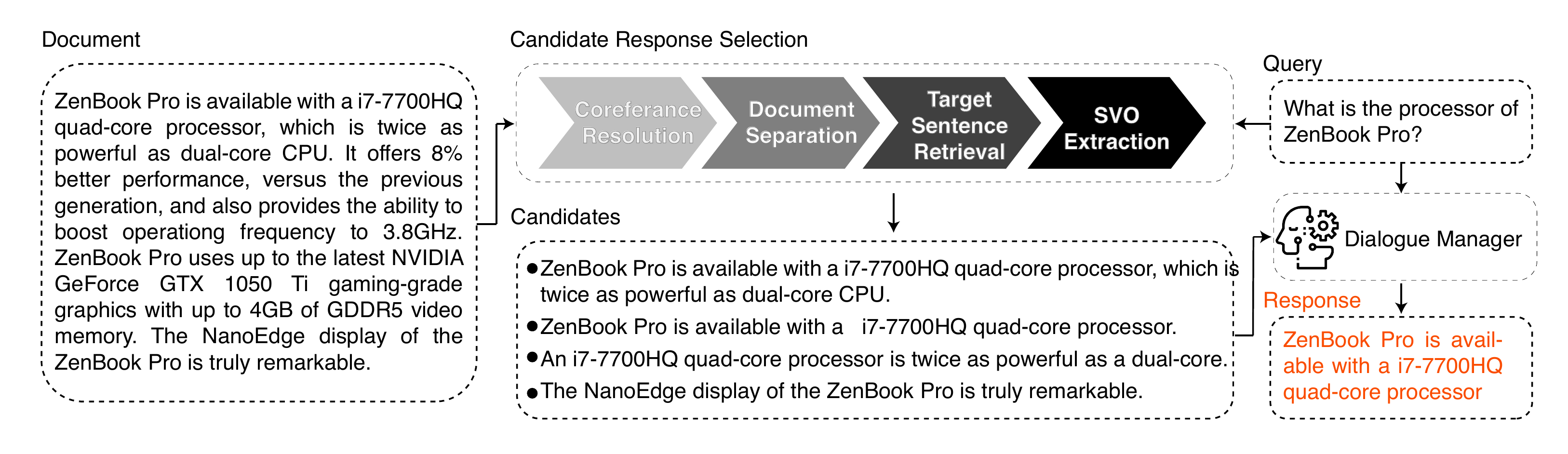}
	\caption{Architecture of Lingke.}
	\label{fig:overview}
\end{figure*}

This section presents the architecture of Lingke, which is overall shown in Figure \ref{fig:overview}. 

The technical components include 1) coreference resolution and document separation, 2) target sentences retrieval, 3) candidate responses generation, followed by a dialouge manager including 4) self-matching attention, 5) response selection and 6) chit-chat response generation. 

The first three steps aim at selecting candidate responses, and in the remaining steps, we utilize sentences from previous conversations to select the most proper response. For multi-turn conversation modeling, we develop a dialogue manager which employs self-matching attention strategy and sequential utterance-response matching to distill pivotal information from the redundant context and determine the most proper response from the candidates.

\paragraph{Coreference Resolution and Document Separation}
Since the response is usually supposed to be concise and coherent, we first separate a given document into sentences. However, long documents commonly involve complex reference relations. A direct segmentation might result in severe information loss. So before the separation, we used \emph{Stanford CoreNLP} \cite{manning-EtAl:2014:P14-5}  \footnote{\url{https://stanfordnlp.github.io/CoreNLP/index.html}} to accomplish the coreference resolution. After the resolution, we cut the document into sentences $A=\{A_{1},A_{2},...,A_{n}\}$. 

\paragraph{Target Sentences Retrieval}
There is abundant information in the whole document, but what current message cares about just exists in some paragraphs or even sentences. So before precise processing, we need to roughly select sentences which are relevant with the current message. We used \emph{Apache Lucene} to accomplish the retrieval. Given sentence collection $A$ from step 1, we retrieve $k$ relevant sentences $ E=\{E_{1},E_{2},...,E_{k}\}$. In our system, the value of $k$ is 2.

\begin{figure}\small\centering
	\begin{minipage}[t]{0.37\linewidth}
		\centering
		\includegraphics[width=1\textwidth]{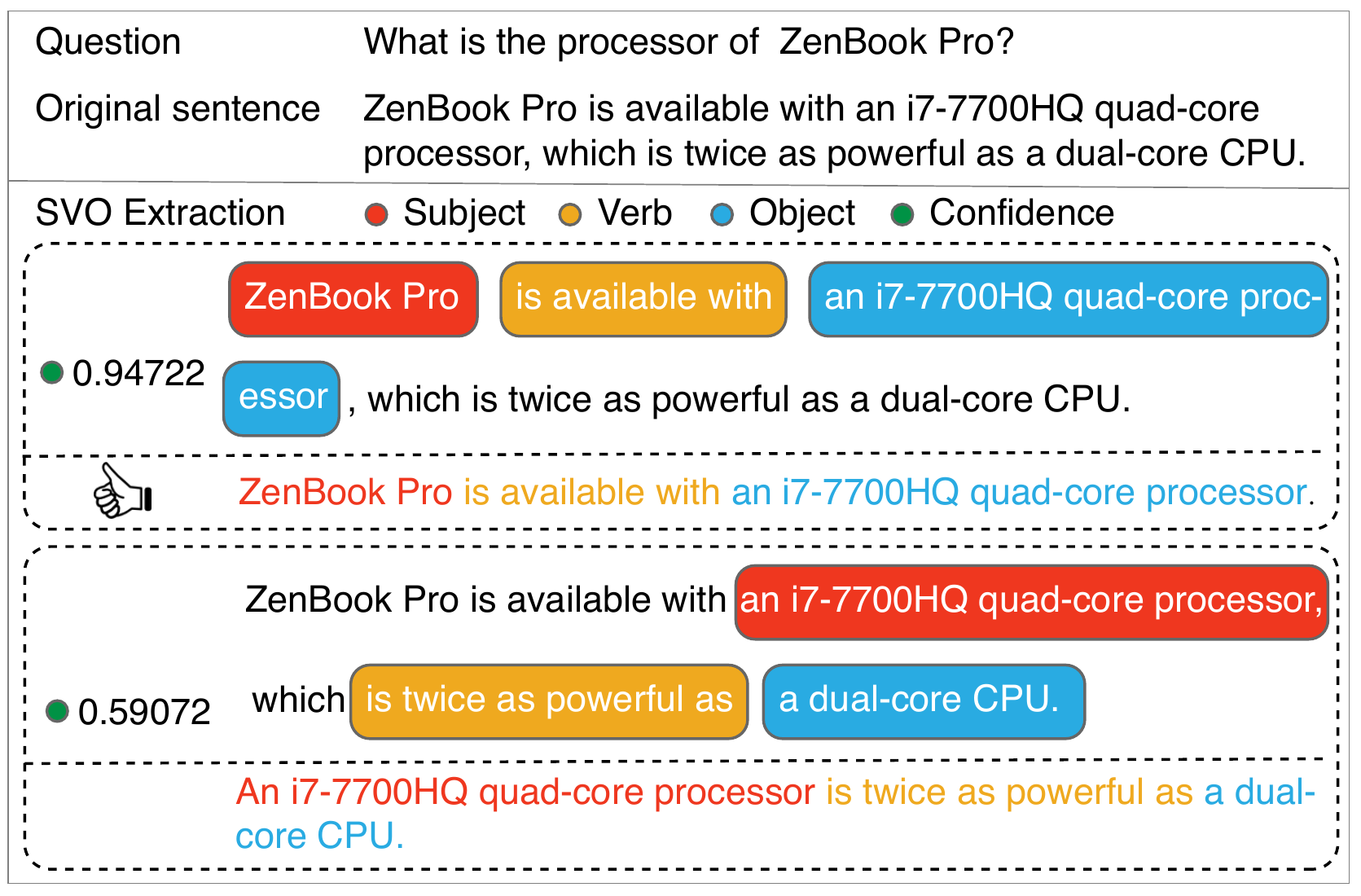}
		\caption{Example of SVO Extraction.}
		\label{tab:example_svo}
	\end{minipage} \quad
	\begin{minipage}[t]{0.60\linewidth}
		\centering
		\includegraphics[width=1\textwidth]{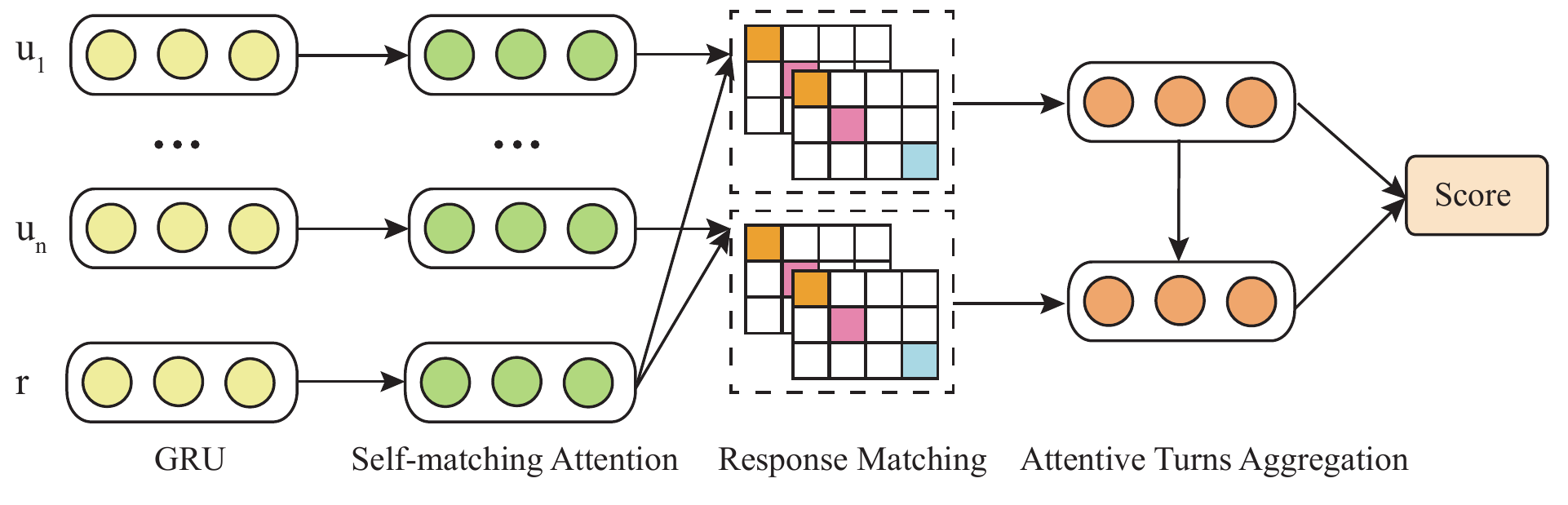}
		\caption{Structure overview of the dialogue manager.}
		\label{fig:dialogue_manager}
	\end{minipage}
\end{figure}

\paragraph{Candidate Responses Generation}
Generally, the response to a conversation can be expressed as a simple sentence, even a few of words. However, sentences from a product introduction document are usually complicated with much information. To extract SVO, we used an open information extraction framework ReVerb \cite {ReVerb2011}, which is able to recognize more than one group of SVO triples (including triples from the clauses). Figure \ref{tab:example_svo} shows an example. Based on an utterance $E_{i}$ from $E$, we extract its SVO triples $E_{s}=\{E_{s1}, E_{s2}, ..., E_{sn}\}$, $E_{v}=\{E_{v1}, E_{v2}, ..., E_{vn}\}$, $E_{o}=\{E_{o1}, E_{o2}, ..., E_{on}\}$, and by concatenating each triple, we obtain multiple of simple sentences $T=\{T_{1}, T_{2}, ..., T_{n}\}$.

The above first three steps generate all sentences and phrases as candidate responses, which are denoted as 
$R=E \cup T $. What we need to do next is to rerank the candidates for the most proper response.

\paragraph{Dialogue Manager}

We combined self-matching attention strategy and sequential utterance-response matching to develop a multi-turn dialogue manager. Figure \ref{fig:dialogue_manager} shows the structure.

\subparagraph{(1) Self-matching Attention}

Since not all of the information is useful, it is a natural thought that adopts self-matching attention strategy to filter redundant information. Before that, we transform raw dialogue data into word embedding \cite {mikolov:2013} firstly. Each conversation utterance or candidate response is fed to the gated recurrent units (GRUs) \cite{Cho2014Learning}. Then, we adopt a self-matching attention strategy \cite {Wang2017Gated} to directly match each utterance or response against itself to distill the pivotal information and filter irrelevant pieces.

\subparagraph{(2) Response Selection}
Following Sequential Matching Network (SMN) \cite {Wu2016Sequential}, we employ sequential matching for multi-turn response selection. Given the candidate response set, it matches each response with the conversation utterances in chronological order and obtains accumulated matching score of the utterance-response pairs to capture significant information and relations among utterances and each candidate response. The one with highest matching score is selected as final response.

\subparagraph{(3) Chit-chat Response Generation}
When given a question irrelevant to current introduction document, \emph{Target Sentences Retrieval} may fail, so we adopt a chit-chat engine to give response when the matching scores of all the candidate responses are below the threshold which is empirically set to 0.3. The chit-chat model is an attention-based seq2seq model \cite {NIPS2014_5346} achieved by a generic deep learning framework \emph{OpenNMT}\footnote{\url{http://opennmt.net/OpenNMT}}. The model is trained on twitter conversation data, which has 370K query-reply pairs, and 300K non-duplicate pairs are selected for training.

\section{Experiment}
\paragraph{Dataset}
We evaluate Lingke on a dataset from our \emph{Taobao}\footnote{It's the largest e-commerce platform in China. \url{https://www.taobao.com}. } partners, which is a collection of conversation records between customers and customer service staffs. It contains over five kinds of conversations, including chit-chat, product and discount consultation, querying delivery progress and after-sales feedback. We converted it into the structured multi-turn format as in \cite{Lowe2015The,Wu2016Sequential}. The training set has 1 million multi-turn dialogues totally, and 10K respectively in validation and test set.

\begin{table}[h]\small
	\centering
	{\small
		\begin{tabular}{l|c|c|c|c|c|c|c|c}
			\hline			\hline
			& TF-IDF  & RNN &CNN & LSTM & BiLSTM & Multi-View & SMN & \textbf{Our model}\\
			\hline
		    ${\rm R_{10} @1} $ 	&  0.159 &  0.325 &0.328 &0.365&  0.355 &  0.421 & 0.453 &\textbf{0.476} \\
		    ${\rm R_{10} @2} $ 	& 0.256  &0.463  &0.515& 0.536 & 0.525 & 0.601 & 0.654&\textbf{0.672} \\
			${\rm R_{10} @5} $  & 0.477 & 0.775 &0.792& 0.828 &  0.825 & 0.861 &0.886& \textbf{0.893}\\
				\hline			\hline
		\end{tabular}
	}
		\tabcaption{\label{tab:result_osc} Comparison of different models.}
\end{table}

\paragraph{Evaluation}

Our model is compared with recent single-turn and multi-turn models, of which the former are in \cite {Kadlec2015Improved,Lowe2015The} including TF-IDF, Convolutional Neural Network (CNN), Recurrent Neural Network (RNN), LSTM and biLSTM. These models concatenate the context utterances together to match a response. Multi-view model \cite {Zhou2016Multi} models utterance relationships from word sequence view and utterance sequence view, and Sequential Matching Network \cite {Wu2016Sequential} matches a response with each utterance in the context. We implemented all the models following the same hyper-parameters from corresponding literatures \cite{Wu2016Sequential,Lowe2015The}. Our evaluation is based on Recall at position $k$ in $n$ candidates ($R_{}n@k$). Results in Table \ref{tab:result_osc} show that our model outperforms all other models, indicating filtering redundant information within utterances could improve the performance and relationships among utterances and response can not be neglected.

\section{Usability and Analysis}

In this section, we will discuss the usability of Lingke. In situation of lacking enough dialogue data such as when a new product is put on an online shop, Lingke only needs an introduction document to respond to customers. Because of the chit-chat response generation engine, Lingke can easily deal with any commodity-independent conversations. Thanks to our multi-turn model, Lingke will not get confused when customer gives incomplete questions which need to be understood based on context.

\begin{figtab}
	\begin{minipage}[b]{0.5\linewidth}
		\centering
		\includegraphics[width=\textwidth]{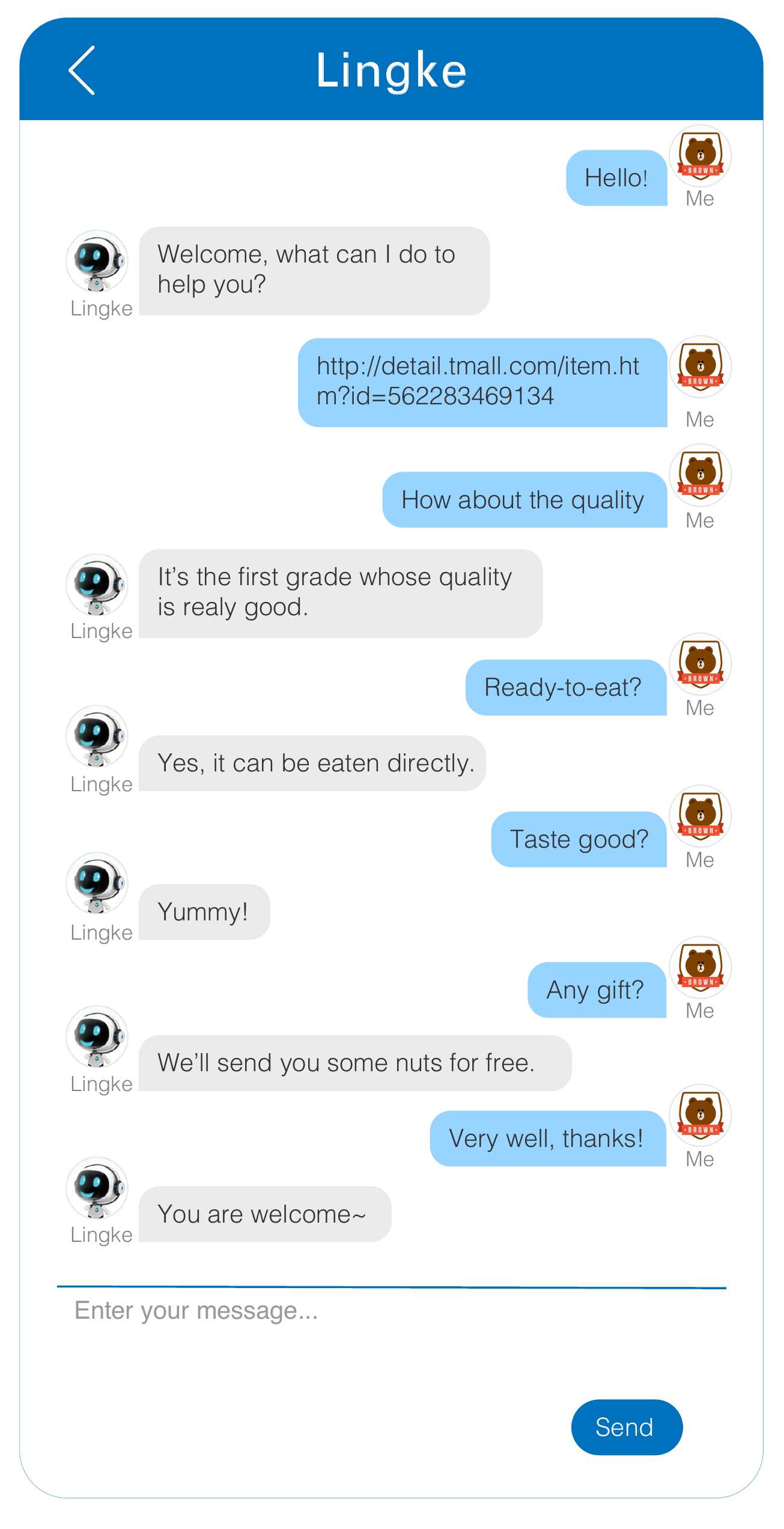}
		\figcaption{A conversation record based example.}\label{fig:shopping}
	\end{minipage}\quad
	\begin{minipage}[b]{0.496\linewidth}
		\centering
		\includegraphics[width=\textwidth]{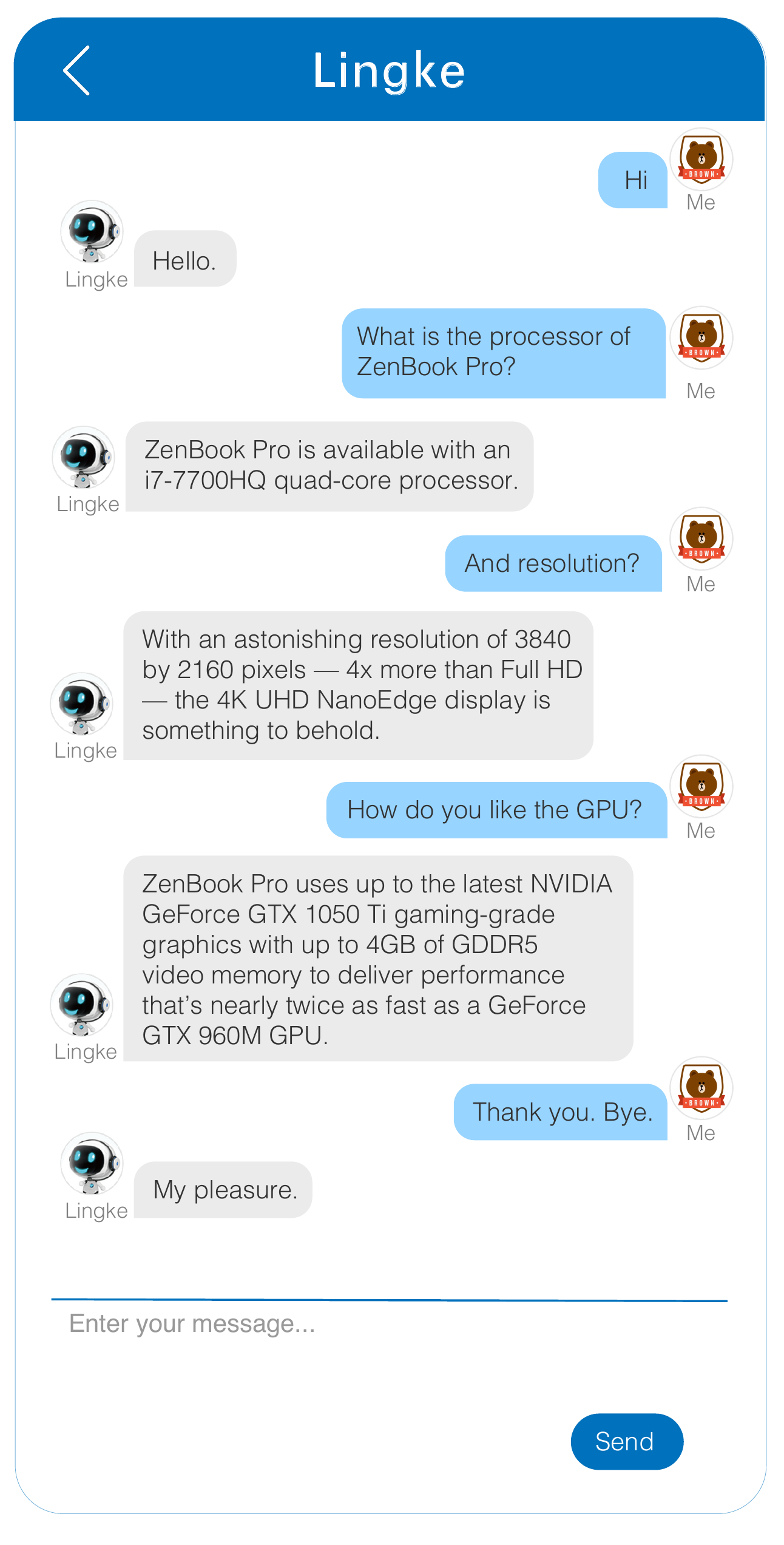}
		\figcaption{A document-based example.}\label{fig:scenario}
	\end{minipage}
\end{figtab}

Figure \ref{fig:shopping}-\ref{fig:scenario} show two typical application scenarios of Lingke, namely, \emph{conversation record based} and \emph{document-based} ones, which vary based on the training corpus. Figure \ref{fig:shopping} shows Linke can effectively respond to the customer shopping consultations. The customer sends a product link and then Lingke recognizes it, and when the customer asks production specifications Lingke will give responses based on information from the context and the conversation record. Figure \ref{fig:scenario} shows a typical scenario when a customer consults Lingke about a new product. The customer starts with a greeting, which is answered by chit-chat engine. Then the customer asks certain features of a product. Note that the second response comes from a sentence which has a redundant clause, and main information the customer cares about has been extracted. In the third user utterance, words like ``What" and ``ZenBook Pro" are omitted, which can be deducted from the prior question. Such pivotal information from the context is distilled and utilized to determine proper response with the merit of self-matching attention and multi-turn modeling.

The user utterances of examples in this paper and our online demo are relatively simple and short, which usually aim at only one feature of the product. In some cases, when the customer utterance becomes more complex, for example, focusing on more than one feature of the product, Lingke may fail to give complete response. A possible solution is to concatenate two relevant candidate responses, but the key to the problem is to determine the intents of the customer.

\section{Conclusion}

We have presented a fine-grained information retrieval augmented chatbot for multi-turn conversations. In this paper, we took e-commerce product introduction as example, but our solution will not be limited to this domain. In our future work, we will add the mechanism of intent detection, and try to find solutions of how to deal with introduction document that contains more than one object.

\bibliography{acl2018}

\begin{thebibliography}{}

\bibitem[\protect\citename{Cho \bgroup et al.\egroup }2014]{Cho2014Learning}
Kyunghyun Cho, Bart~Van Merrienboer, Caglar Gulcehre, Dzmitry Bahdanau, Fethi
  Bougares, Holger Schwenk, and Yoshua Bengio.
\newblock 2014.
\newblock Learning phrase representations using {RNN} encoder-decoder for
  statistical machine translation.
\newblock In {\em EMNLP 2014}, pages 1724--1734.

\bibitem[\protect\citename{Cui \bgroup et al.\egroup }2017]{Cui2017SuperAgent}
Lei Cui, Shaohan Huang, Furu Wei, Chuanqi Tan, Chaoqun Duan, and Ming Zhou.
\newblock 2017.
\newblock Superagent: A customer service chatbot for e-commerce websites.
\newblock In {\em ACL 2017, Demo}, pages 97--102.

\bibitem[\protect\citename{Fader \bgroup et al.\egroup }2011]{ReVerb2011}
Anthony Fader, Stephen Soderland, and Oren Etzioni.
\newblock 2011.
\newblock Identifying relations for open information extraction.
\newblock In {\em EMNLP 2011}, Edinburgh, Scotland, UK, July 27-31.

\bibitem[\protect\citename{Huang \bgroup et al.\egroup }2018]{Huang2018Moon}
Yafang Huang, Zuchao Li, Zhuosheng Zhang, and Hai Zhao.
\newblock 2018.
\newblock {Moon IME:} neural-based chinese pinyin aided input method with
  customizable association.
\newblock In {\em Proceedings of the 56th Annual Meeting of the Association for
  Computational Linguistics (ACL 2018), System Demonstration}.

\bibitem[\protect\citename{Kadlec \bgroup et al.\egroup
  }2015]{Kadlec2015Improved}
Rudolf Kadlec, Martin Schmid, and Jan Kleindienst.
\newblock 2015.
\newblock Improved deep learning baselines for {Ubuntu} corpus dialogs.
\newblock {\em arXiv preprint arXiv:1510.03753}.

\bibitem[\protect\citename{Lowe \bgroup et al.\egroup }2015]{Lowe2015The}
Ryan Lowe, Nissan Pow, Iulian Serban, and Joelle Pineau.
\newblock 2015.
\newblock {The Ubuntu Dialogue Corpus}: A large dataset for research in
  unstructured multi-turn dialogue systems.
\newblock In {\em SIGDIAL 2015}, pages 285--294.

\bibitem[\protect\citename{Manning \bgroup et al.\egroup
  }2014]{manning-EtAl:2014:P14-5}
Christopher~D. Manning, Mihai Surdeanu, John Bauer, Jenny Finkel, Steven~J.
  Bethard, and David McClosky.
\newblock 2014.
\newblock The {Stanford} {CoreNLP} natural language processing toolkit.
\newblock In {\em ACL 2014, Demo}, pages 55--60.

\bibitem[\protect\citename{Mikolov \bgroup et al.\egroup }2013]{mikolov:2013}
Tomas Mikolov, Kai Chen, Greg Corrado, and Jeffrey Dean.
\newblock 2013.
\newblock Efficient estimation of word representations in vector space.
\newblock {\em arXiv preprint arXiv:1301.3781}.

\bibitem[\protect\citename{Qiu \bgroup et al.\egroup }2017]{Qiu2017AliMe}
Minghui Qiu, Feng~Lin Li, Siyu Wang, Xing Gao, Yan Chen, Weipeng Zhao, Haiqing
  Chen, Jun Huang, and Wei Chu.
\newblock 2017.
\newblock Alime chat: A sequence to sequence and rerank based chatbot engine.
\newblock In {\em ACL 2017}, pages 498--503.

\bibitem[\protect\citename{Sutskever \bgroup et al.\egroup
  }2014]{NIPS2014_5346}
Ilya Sutskever, Oriol Vinyals, and Quoc~V Le.
\newblock 2014.
\newblock Sequence to sequence learning with neural networks.
\newblock In Z.~Ghahramani, M.~Welling, C.~Cortes, N.~D. Lawrence, and K.~Q.
  Weinberger, editors, {\em Advances in Neural Information Processing Systems
  27}, pages 3104--3112. Curran Associates, Inc.

\bibitem[\protect\citename{Wang \bgroup et al.\egroup }2017]{Wang2017Gated}
Wenhui Wang, Nan Yang, Furu Wei, Baobao Chang, and Ming Zhou.
\newblock 2017.
\newblock Gated self-matching networks for reading comprehension and question
  answering.
\newblock In {\em ACL 2017}, pages 189--198.

\bibitem[\protect\citename{Wu \bgroup et al.\egroup }2017]{Wu2016Sequential}
Yu~Wu, Wei Wu, Chen Xing, Ming Zhou, and Zhoujun Li.
\newblock 2017.
\newblock Sequential matching network: A new architecture for multi-turn
  response selection in retrieval-based chatbots.
\newblock In {\em ACL 2017}, pages 496--–505.

\bibitem[\protect\citename{Yan \bgroup et al.\egroup }2016a]{Yan2016DocChat}
Zhao Yan, Nan Duan, Junwei Bao, Peng Chen, Ming Zhou, Zhoujun Li, and Jianshe
  Zhou.
\newblock 2016a.
\newblock Docchat: An information retrieval approach for chatbot engines using
  unstructured documents.
\newblock In {\em ACL 2016}, pages 516--525.

\bibitem[\protect\citename{Yan \bgroup et al.\egroup }2016b]{Yan2017Building}
Zhao Yan, Nan Duan, Peng Chen, Ming Zhou, Jianshe Zhou, and Zhoujun Li.
\newblock 2016b.
\newblock Building task-oriented dialogue systems for online shopping.
\newblock In {\em AAAI-17}, pages 516--525.

\bibitem[\protect\citename{Zhang \bgroup et al.\egroup }2017]{Zhang2017Benben}
Wei~Nan Zhang, Ting Liu, Bing Qin, Yu~Zhang, Wanxiang Che, Yanyan Zhao, and
  Xiao Ding.
\newblock 2017.
\newblock Benben: A {Chinese} intelligent conversational robot.
\newblock In {\em ACL 2017, Demo}, pages 13--18.

\bibitem[\protect\citename{Zhang \bgroup et al.\egroup }2018a]{zhang2018SubMRC}
Zhuosheng Zhang, Yafang Huang, and Hai Zhao.
\newblock 2018a.
\newblock Subword-augmented embedding for cloze reading comprehension.
\newblock In {\em Proceedings of the 27th International Conference on
  Computational Linguistics (COLING 2018)}.

\bibitem[\protect\citename{Zhang \bgroup et al.\egroup }2018b]{zhang2018dua}
Zhuosheng Zhang, Jiangtong Li, Pengfei Zhu, and Hai Zhao.
\newblock 2018b.
\newblock Modeling multi-turn conversation with deep utterance aggregation.
\newblock In {\em Proceedings of the 27th International Conference on
  Computational Linguistics (COLING 2018)}.

\bibitem[\protect\citename{Zhou \bgroup et al.\egroup }2016]{Zhou2016Multi}
Xiangyang Zhou, Daxiang Dong, Hua Wu, Shiqi Zhao, Dianhai Yu, Hao Tian, Xuan
  Liu, and Rui Yan.
\newblock 2016.
\newblock Multi-view response selection for human-computer conversation.
\newblock In {\em EMNLP 2016}, pages 372--381.

\end{thebibliography}
\bibliographystyle{acl}

\end{document}